\newif\ifcomments  %
\newif\ifsp %

\commentstrue

\documentclass[12pt, twocolumn]{IEEEtran}
\usepackage{amsthm,amsmath}
\usepackage{bbm}
\usepackage{times}
\usepackage{array,float}
\usepackage{url}
\usepackage{amstext,amssymb}
\usepackage{hyphenat}
\usepackage{verbatim}
\usepackage{dsfont}
\usepackage{bm}
\usepackage{wrapfig}
\usepackage[noend]{algorithmic}
\usepackage{algorithm}
\usepackage{graphicx,color}
\usepackage{xparse,etoolbox}
\usepackage{threeparttable}
\usepackage{dsfont}
\usepackage{xspace}
\usepackage[utf8]{inputenc}
\usepackage{outlines}
\usepackage{comment}
\usepackage{footnote}
\usepackage{multirow}
\usepackage{ulem}
\usepackage{cleveref}
\usepackage{subcaption}
\usepackage{authblk}
\usepackage{booktabs}

\newtheorem{lemma}{Lemma}[section]

\newtheorem{defn}[lemma]{Definition}

\renewcommand{\paragraph}[1]{\vspace{3pt}\noindent\textbf{#1}}

\newcommand{\norm}[1]{\|#1\|}

\newcommand{\eps}{\epsilon}

\newcommand{\cA}{\mathcal{A}}

\newcommand{\cS}{\mathcal{S}}

\newcommand{\R}{\mathcal{R}}

\newcommand{\grad}{\bigtriangledown}
\newcommand{\mypar}[1]{\bigskip
	\noindent{\textbf{\em {#1}:}}}

\newcommand{\ignore}[1]{}

\definecolor{darkgreen}{rgb}{0,0.4,0.0}

\makeatletter
\def\blfootnote{\gdef\@thefnmark{}\@footnotetext}
\makeatother

\newcommand{\SUB}[1]{\hspace{-0.15in} \textbf{#1}}

\newcommand{\moments}{\mathcal{M}}
\newcommand{\sample}{\mathcal{C}}
\newcommand{\update}{\Delta}
\newcommand{\Normal}{\mathcal{N}}
\newcommand{\lepochs}{\ensuremath{E}}

\newcommand{\lbs}{\ensuremath{B}}
\newcommand{\loss}{\ell}

\makeatletter
\newcommand{\skipitems}[1]{\addtocounter{\@enumctr}{#1}}
\makeatother

\renewcommand{\paragraph}[1]{\medskip
\noindent{\bf{#1}}}

\newcommand{\mycaptionof}[2]{\captionof{#1}{#2}}

\begin{document}
\title{Training Production Language Models \\ without Memorizing User Data}
\ifsp
\author{Anonymous authors}
\else
\author{Swaroop Ramaswamy\textsuperscript{*}\thanks{\textsuperscript{*}Equal contribution}}
\author{Om Thakkar\textsuperscript{*}}
\author{Rajiv Mathews}
\author{Galen Andrew}
\author{H. Brendan McMahan}
\author{Françoise Beaufays}
\affil{Google LLC,\\
Mountain View, CA, U.S.A. \\
\texttt{\{swaroopram, omthkkr, mathews, galenandrew, mcmahan, fsb\}   @google.com}}
\fi
\maketitle

\begin{abstract}
 This paper presents the first consumer-scale next-word prediction (NWP) model trained with Federated Learning (FL) while leveraging the Differentially Private Federated Averaging (DP-FedAvg) technique.
 There has been prior work on building practical FL infrastructure, including work demonstrating the feasibility of training language models on mobile devices using such infrastructure. It has also been shown (in simulations on a public corpus) that it is possible to train NWP models with user-level differential privacy using the DP-FedAvg algorithm. Nevertheless, training production-quality NWP models with DP-FedAvg in a real-world production environment on a heterogeneous fleet of mobile phones requires addressing numerous challenges. For instance, the coordinating central server has to keep track of the devices available at the start of each round and sample devices uniformly at random from them, while ensuring \emph{secrecy of the sample}, etc.   Unlike all prior privacy-focused FL work of which we are aware, for the first time we demonstrate the deployment of a differentially private mechanism for the training of a production neural network in FL, as well as the instrumentation of the production training infrastructure to perform an end-to-end empirical measurement of unintended memorization.
\end{abstract}

\section{Introduction}

Next word prediction (NWP) is the task of providing the most probable next word or phrase given a small amount of preceding text.
 \ifsp X\footnote{Anonymized for blind review.} \else Gboard \fi  is a virtual keyboard for touchscreen mobile devices that provides features such as auto-correction and word completion, in addition to next-word prediction.
Trained language models (LMs) are used to perform the task of NWP on user-generated data. To provide high utility, they are trained using user-generated data as well. However, such data can be privacy sensitive; it can include chats, text messages, and search queries. Federated learning~\cite{FL1,AOPFL} is a distributed learning approach that enables training models without the need to centralize user data.
There has been work~\cite{FL19} in developing a scalable production system for FL, based on TensorFlow~\cite{tf}, in the domain of mobile devices. Recent work~\cite{NWP18} has used this system to train a model for the NWP task. In this work, we build on the approach of \cite{NWP18}.

In this work, our primary goal is the protection of private user data from an adversary with access to the final machine learning model trained on user data via FL; we thus assume the server implementing FL is trusted. Since such models are typically deployed to many millions of devices for on-device inference, access to the model and its predictions cannot realistically be controlled. Thus, ensuring private information cannot be extracted from the model is essential.
Providing such guarantees with weaker trust assumptions for the server (honest-but-curious, or malicious) is a valuable goal, but it requires different techniques and is beyond the scope of this work~\cite{AOPFL}.

Differential privacy (DP)~\cite{DMNS06, DKMMN06} provides a gold standard for performing learning tasks over sensitive data. Intuitively, DP prevents an adversary from confidently making any conclusions about whether any particular data record was used in training a model, even while having access to the model and arbitrary external side information. For machine learning, two granularities of a data record are particularly relevant,  \emph{example-level}, and \emph{user-level} (though notions in between these have been considered, for example ``element-level" \cite{asi2019element}).
Many prior works in DP machine learning (ML) \cite{CMS11, BassilyST14, DPDL, papernot2017semi, wu2017bolt, papernot2018scalable, INSTTW19} deal with example-level privacy, i.e., providing privacy guarantees for any single example in a dataset. However, in tasks like language modeling, such a guarantee can be quite weak, as any individual user may contribute thousands of examples to the training corpus. FL is naturally suited to the strictly stronger notion of user-level privacy (\cite{MRTZ18, JTT18, DPFGAN19, TAM19}), which provides guarantees for all the examples contributed by any individual user in the training process.
Differential privacy comprises two main components. First, a \emph{DP mechanism} is a randomized procedure where typically 1) an upper bound on the sensitivity of the mechanism to any one user's data is enforced, and 2) noise calibrated to that sensitivity is added to the output. We deploy such a mechanism (see Section~\ref{sec:dpfa} for more details).
Second, such a mechanism is accompanied by a formal DP guarantee characterized by two parameters $\eps$ and $\delta$ that upper-bound the privacy loss of the mechanism.

Prior work \cite{MRTZ18} provides a technique, called Differentially Private Federated Averaging (DP-FedAvg), for training neural networks (including recurrent language models) with user-level DP via FL. It has shown that good privacy-utility trade-offs are possible in idealized simulated FL environments with a large number of users. Federated learning alone offers direct privacy benefits by keeping data decentralized, allowing client devices to control their participation, aggregating early, and only sending focused ephemeral updates to the server. One of the contributions of this work is highlighting that, perhaps surprisingly, these very privacy benefits of FL make it \emph{more challenging} for the server to provide a proof of a specific $(\eps, \delta)$-DP guarantee since it has limited visibility and control of the overall decentralized training mechanism. In fact, in production FL systems, the assumptions required by known DP theorems~\cite{BassilyST14, DPDL, MRTZ18} may only hold approximately, or otherwise be difficult to verify. Designing new DP mechanisms and analysis that address these challenges and hence apply to real-world deployments of FL is an important active area for research~\cite{BKMTT}, but in this work we take a complimentary approach.

We deploy the DP-FedAvg mechanism in a real-world system, and then, rather than focusing on proving upper-bounds on $(\eps, \delta)$-DP (which exist, but may be hard for the server to certify), we assess the privacy of our training method using an end-to-end measurement process.  Our evaluation of privacy is based on the \emph{Secret Sharer} framework \cite{CLKES18} (more details in Section~\ref{sec:ss}) for an FL setting~\cite{TRMB}, which can measure \emph{unintended memorization} of user data.
Prior work~\cite{TRMB} has shown via simulations that training generative models with DP-FedAvg does not exhibit such memorization for thousands insertions of \emph{out-of-distribution} phrases in the training data. Our results are noteworthy as our  models are trained in a production setting using actual user data, and are able to tolerate thousands of insertions of out-of-distribution phrases as well, while at the same time providing better utility than the existing benchmark.
We perform this validation as part of a multi-faceted approach to private training, including other techniques like using a fixed vocabulary for the training data, and limiting the number of training data for each individual user.

Even as the theory in differentially private ML advances \cite{CWH, STT}, we believe the end-to-end approach described here will continue to be a vital component of applied ML on private data. A theoretical result applies to an algorithm operating under particular assumptions. In contrast, an end-to-end measurement approach tests a complete software system running under real-world conditions, allowing for instance, the detection of bugs or violated assumptions that would fall outside the scope of theory.

\section{Preliminaries}
\label{sec:prelims}

 \subsection{DP Federated Averaging with Fixed-size Rounds}
 \label{sec:dpfa}

We now present the DP mechanism (Algorithm~\ref{alg.fedavg}) that we employ to train our language model. It closely follows the DP-FedAvg technique in \cite{MRTZ18}, in that per-user updates are clipped to have a bounded $L_2$ norm, and calibrated Gaussian noise is added to the weighted average update to be used for computing the model to be sent in the next round. A slight difference between the DP-FedAvg algorithm in \cite{MRTZ18} and our approach is the way in which client devices are sampled to participate in a given federated round of computation. DP-FedAvg uses Poisson sampling, where for each round, each user is selected independently with a fixed probability. In this work (also, following \cite{DPFGAN19, TRMB}), we instead use fixed-size federated rounds, where a fixed number of users is randomly sampled to participate in each round. A pseudo-code for our mechanism is given in  Algorithm~\ref{alg.fedavg}.

Proving a formal DP guarantee for Algorithm~\ref{alg.fedavg} requires several assumptions like knowledge of the size of the participating user population ($N$), and the server being able to sample uniformly at random among them at each iteration.
Such assumptions  may not always hold in real-world deployments. Section~\ref{sec:prac} presents a detailed discussion of practical considerations for privacy guarantees in real-world FL systems.

\begin{figure}
\centering
\rule{0.49\textwidth}{\heavyrulewidth}
\begin{minipage}[t]{0.49\textwidth}
\begin{center}
\begin{algorithmic}
\SUB{Main training loop:}
   \STATE \emph{parameters:}  round participation fraction $q \in (0, 1]$,  total user population $D$ of size $N \in \mathbb{N}$, noise scale $z \in {\R}^+$, clip parameter $S \in {\R}^+$, total rounds $T$
   \STATE
   \STATE Initialize model $\theta^0$, moments accountant $\moments$
   \STATE Set noise standard deviation $\sigma = \frac{\displaystyle z S}{\displaystyle qN}$
   \FOR{each round $t = 0, 1, 2, \ldots, T$}
     \STATE $\sample^t \leftarrow$ (sample without replacement $q N$ users from population)
     \FOR{each user $k \in \sample^t$ \textbf{in parallel}}
       \STATE $\update^{t+1}_k \leftarrow \text{UserUpdate}(k, \theta^t)$
     \ENDFOR
     \STATE $\update^{t+1} = \frac{\displaystyle 1}{\displaystyle qN} \sum\limits_{k\in \sample^t} \update^{t+1}_k$
     \STATE $\theta^{t+1} \leftarrow \theta^t + \update^{t+1} + \Normal(0, I\sigma^2)$
   \ENDFOR
\end{algorithmic}
\end{center}
\vfill
\end{minipage}
\hfill
\begin{minipage}[t]{0.49\textwidth}
\begin{center}
\begin{algorithmic}
 \SUB{UserUpdate($k, \theta^0$):}
  \STATE \emph{parameters:}  number of local epochs $\lepochs \in \mathbb{N}$, batch size $\lbs \in \mathbb{N}$, learning rate $\eta \in {\R}^+$, clip parameter $S \in {\R}^+$, loss function $\loss(\theta; b)$
  \STATE
  \STATE $\theta \leftarrow \theta^0$
  \FOR{each local epoch $i$ from $1$ to $\lepochs$}
    \STATE $\mathcal{B} \leftarrow$ ($k$'s data split into size $\lbs$ batches)
    \FOR{each batch $b \in \mathcal{B}$}
      \STATE $\theta \leftarrow \theta - \eta \grad \loss(\theta; b)$
    \ENDFOR
 \ENDFOR
 \STATE $\update = \theta - \theta^0$
 \STATE return update $\update_k = \update \cdot \min\left(1, \frac{S}{\norm{\update}}\right)$ \hfill \emph{// Clip}

\end{algorithmic}
\end{center}
\vfill
\end{minipage}
\rule{0.49\textwidth}{\heavyrulewidth}

\mycaptionof{algorithm}{DP-FedAvg with fixed-size federated rounds, used to train our language model.}\label{alg.fedavg}
\end{figure}

\subsection{Measuring Unintended Memorization}
\label{sec:ss}

 We use the Secret Sharer technique from \cite{CLKES18} as a proxy for measuring how much private information might be extracted from such a model. Our approach is designed to over-estimate what a realistic adversary might learn (more details in Section~\ref{sec:exp_set}).  However, unlike a formal DP guarantee, this empirical approach cannot rule out the possibility that some more clever technique (for example, one that directly inspects the model parameters) might reveal more. Thus, developing more sophisticated attacks (memorization measurement techniques) is an important complimentary line of research.

Now, we describe the Secret Sharer framework. First,  random sequences called canaries are inserted into the training data. The canaries are constructed based on a prefixed format sequence. For instance, to design the framework for a character-level model, the format could be ``My SSN is $xxx$-$xx$-$xxxx$", where each $x$ can take a random value from digits 0 to 9. Next, the target model is trained on the modified dataset containing the canaries. Lastly, methods like Random Sampling and Beam Search (both formally defined in Section~\ref{sec:mem}) are used to efficiently measure the extent to which the model has ``memorized" the inserted random canaries, and whether it is possible for an adversary with partial knowledge to extract the canary.
For instance, if a canary is classified as memorized via our Random Sampling method, then an adversary with a ``guess" of the canary can be confident with very high probability whether the guess is correct just by randomly sampling other phrases and evaluating their perplexities on the given model.

\section{Implementation Details}
\label{sec:impl}
 In this section, we start by providing the details of our implementation, and state the performance of our NWP model. We show that even with clipping client updates and a large amount of noise addition, our NWP model has superior utility than the existing baseline n-gram Finite State Transducer (FST) model. The FST model is a Katz-smoothed Bayesian interpolated LM that is augmented with other smaller LMs such as a user history LM.

\subsection{Model Architecture and Hyperparameters}

The model architecture we use mirrors the one used in \cite{NWP18}. We use a single layer CIFG-LSTM~\cite{cifg} neural network with shared weights between the input embedding layer and the output projection layer. The overall number of parameters in the model is 1.3M.

Typically, tuning hyperparameters for neural networks requires training several models with various hyperparameter settings. Instead of tuning hyperparameters on sensitive user data, we tune the hyperparameters by training the same model with DP-FedAvg on a public dataset, namely the Stack Overflow corpus.\footnote{\url{https://www.tensorflow.org/federated/api_docs/python/tff/simulation/datasets/stackoverflow/load_data}} By tuning hyperparameters on a public dataset, we avoid incurring any additional privacy cost.

When training on real devices, we use the hyperparameters that performed best on the Stack Overflow dataset. The only change we make is to the words in the vocabulary; when training on real devices we train on only devices containing Spanish language data.

For all hyperparameter tuning, we train models with 500 users participating in every round, and add Gaussian noise with $\sigma=3.2\times10^{-5}$ to the average of their clipped updates. Note that to get any actual privacy guarantees, we would have to train models with a significantly larger number of users participating per round for the same amount of noise added ($\sigma$). Since we are doing our hyperparameter tuning on a public dataset, we are only interested in the utility characteristics of the trained models, not any privacy guarantees.

We evaluate the performance of all models on the recall metric, defined as the ratio of the number of correct predictions to the total number of words. Recall for the highest-likelihood candidate (top-1 recall) is important for \ifsp X\else Gboard\fi as these are presented in the center of the suggestion strip where users are more likely to see them. Since \ifsp X\else Gboard\fi includes multiple candidates in the suggestion strip, top-3 recall is also of interest.

The best performing model hyperparameters on the Stack Overflow dataset are listed in Table~\ref{table:best_hyperparameters}. We also run a few ablation studies to study the effect of various hyperparameters on recall. We find that using momentum as the server optimizer and clipping around 90\% of the clients per round gives best results. We also find that the utility is not affected by different choices of client batch sizes. Refer to Appendix~\ref{sec:abl} for more details on the ablation studies.

\begin{table}[h!]
    \centering
    \begin{tabular}{ l  c }
     \toprule
     Hyperparameter &  Value \\
     \midrule
     Server optimizer & Momentum \\
     Server learning rate ($\eta_s$) & 1.0 \\
     Server momentum ($\mu$) & 0.99 \\
     Client batch size ($|b|$) & 50 \\
     Client learning rate ($\eta_c$) & 0.5 \\
     Clipping norm ($S$) & 0.8 \\
     \bottomrule
    \end{tabular}
    \caption{Hyperparameter values for the best performing model configuration on Stack Overflow.}
    \label{table:best_hyperparameters}
\end{table}

\subsection{Production Training}
\label{sec:prod}
We train a model using the DP-FedAvg algorithm on real devices running \ifsp X\else Gboard\fi, with the model configuration specified in Table~\ref{table:best_hyperparameters}. We aggregate updates from 20000 clients on each round of training, and add Gaussian noise with standard deviation $\sigma=3.2\times10^{-5}$ to the average of their clipped updates. The model converges after $T=2000$ rounds of training, which took about three weeks to complete.

\subsection{Live Experiments}
\newcommand{\T}{\rule{0pt}{4ex}}
\begin{table}[h!]
    \renewcommand{\arraystretch}{1.1}
    \centering
    \begin{tabular}{ l c c c c}
     \toprule
     Metric & N-gram FST  & Our NWP model & Relative  \\
     & (Baseline) & [This paper]  & Change (\%) \\
     \midrule
     Top-1 Recall & $10.24$ & $11.03$ & $+7.77\%$ \\
     & & & $(7.49, 8.06)$ \\
     \T Top-3 Recall & $18.09$ & $19.25$ & $+6.40\%$\\
     & & & $(6.17, 6.63)$ \\
     \T CTR & $1.84$ & $1.92$ & $+4.31\%$\\
     & & & $(2.17, 6.45)$ \\
     \bottomrule
    \end{tabular}
    \caption{Live inference experiment results.}
    \label{table:LiveExpResults}
\end{table}

We compare the results from our model with the baseline n-gram FST model in a live experiment. In addition to top-1 recall and top-3 recall, we also look at the prediction click-through rate metric, defined as the ratio of number of clicks on prediction candidates to the number of proposed prediction candidates.

The top-1 recall and top-3 recall in this experiment are measured over the number of times users are shown prediction candidates.
The prediction click-through rate (CTR) is defined as the ratio of the number of clicks on prediction candidates to the number of proposed prediction candidates.
Quoted 95\% confidence interval errors for all results are derived using the jackknife method with user buckets. Table~\ref{table:LiveExpResults} summarizes the recall and CTR metrics in live experiment for our NWP model trained using DP-FedAvg, and the baseline n-gram FST model.

The live experiment results from Table~\ref{table:LiveExpResults} show that the NWP model significantly outperforms the baseline n-gram FST, in both recall and CTR metrics.
This is consistent with the observations from \cite{NWP18}. These gains are impressive given that the n-gram model FST includes personalized components such as user history.

\section{Evaluating for Unintended Memorization}
\label{sec:mem}

There is a growing line of work (\cite{mod-inv1, mod-inv2,mem-inf1,CLKES18,SS19, TRMB}) demonstrating that neural networks can leak information about their underlying training data in many ways. Given that we train next-word prediction models in this work, we focus on the Secret Sharer frameworks from \cite{CLKES18, TRMB} designed to measure the resilience of generative models obtained via a training procedure, against the unintended memorization of rarely-occurring phrases in a dataset. Specifically, we extend the idea of the Federated Secret Sharer~\cite{TRMB}, which focused on user-based datasets that are typical in FL, to a production setting. Through an extensive empirical evaluation, we demonstrate the remarkable extent to which training models via our implementation is able to withstand such memorization.

\subsection{Experiment Setup}
\label{sec:exp_set}
Next, we describe the setup of our empirical evaluation. In the following, we detail the various stages of our procedure, including creating \emph{secret-sharing} synthetic devices, construction of the canaries added into the synthetic devices, insertion of the synthetic devices into our FL training procedure, and the techniques used for measuring unintended memorization of a generative model.

 \paragraph{Network architecture, and training corpus:} Since we want to measure memorization for the models trained via our implementation, we start with the same network architecture and training corpus as described in Section~\ref{sec:impl} for conducting the experiments in this section.

     \paragraph{Canary construction:} We opt for inserting five-word canaries as our model is not efficient at encoding longer contexts.
    Each word in a canary is chosen uniformly at random (u.a.r.) from the 10K model vocabulary.
    It is important to note that we want to measure \emph{unintended} memorization for our models, i.e., memorization of out-of-distribution phrases, which is in fact orthogonal to our learning task.
    Hence, to be able to obtain such phrases with very high probability, our canaries are constructed using randomly sampled words.  For instance, our inserted canaries consist of phrases like ``extranjera conciertos mercadeo cucharadas segundos", ``domicilio mariposa haberlo cercanas partido", ``ve trabajador corrida sabemos cuotas", etc.

 \paragraph{Secret-sharing synthetic devices:}
 Since our models involve training on actual devices, we create various synthetic devices containing canaries in their training data, and have them participate in the training along with actual devices.
 To make this setting more realistic, the synthetic devices contain sentences from a public corpus in addition to the canaries.
 Each canary is parameterized by two parameters, $n_u$ and $n_e$. The number of synthetic devices sharing the canary is denoted by $n_u$. Each such synthetic device contains $n_e$ copies of the canary, and $(200 - n_e)$ sentences randomly sampled from the public corpus.
 We consider canaries with configurations in the cross product of $n_u \in \{1, 4, 16\}$ and $n_e \in \{1, 14, 200  \}$, and we have three different canaries for each $(n_u, n_e)$ configuration. These parameters result in the insertion of 27 different canaries, and a total of $3 \cdot 3 \cdot (1 + 4 + 16) = 189$ unique synthetic devices participating in the training process. We avoid adding more than three different canaries for each $(n_u, n_e)$ configuration so as to not overwhelm the training data with canaries.

 \paragraph{Training procedure:}
    We use the training procedure described in Section~\ref{sec:impl} for training our models, with the only difference being that for each round of training, we include all the secret-sharing synthetic devices to be available for being sampled.
    The rate of participation of the synthetic devices is 1-2 orders of magnitude higher than any actual device due to two main factors. First, our synthetic devices are available throughout the training process, which is not the case for actual devices. Moreover, even when the actual devices are available, their participation in the training process is coordinated by our load-scheduling mechanism called Pace Steering~\cite{FL19}, which lowers the next scheduling priority of a device once it has participated in training (to restrict multiple participations within any short phase of training). On the other hand, our synthetic devices don't adhere to Pace Steering, resulting in a further increase in their participation rate.
    Table~\ref{table:mem_seen} shows for each canary configuration, the number of times a canary is encountered by a model trained in our setup.
    From the $(n_u=1, n_e=1)$ configuration, it is easy to see that each secret-sharing synthetic device (for any canary configuration) participates in expectation 1150 times during 2000 rounds of training.
    Note that this should, if anything, increase the chance that a canary phrase will be memorized.

    \begin{table}[h!]
    \centering
    \begin{tabular}{r  r  r}
     \toprule
     $n_u$ & $n_e$ &  Expected \# times  \\
     & & canary seen in training \\
     \midrule
     1 & 1   &  $1,150$ \\
     1 & 14  &  $16,100$ \\
     1 & 200  &  $230,000$ \\
     4 & 1  &  $4,600$ \\
     4 & 14  & $64,400$ \\
     4 & 200  & $920,000$ \\
     16 & 1  &  $18,400$ \\
     16 & 14  &  $257,600$ \\
     16 & 200  &  $3,680,000$ \\
     \bottomrule
    \end{tabular}
    \caption{Expected number of times canaries for each $(n_u, n_e)$ configuration encountered by a model trained in our setup.}
    \label{table:mem_seen}
    \end{table}

  \paragraph{Evaluation methods:} For our evaluation, we denote an inserted canary by $c=(p|s)$, where $p$ is a 2-word prefix, and $s$ is the remaining 3-word sequence. We use the two methods of evaluation used in \cite{TRMB}, namely Random Sampling and Beam Search, to determine if given the canary prefix $p$, the remaining sequence $s$ has been unintentionally memorized by a model.
  \begin{enumerate}
      \item \mypar{Random Sampling (RS) \cite{CLKES18}} First, we define the log-perplexity of a model $\theta$ on a sequence $s = s_1, \ldots, s_n$ given context $p$ as $P_\theta(s | p) = \sum\limits_{i=1}^n \left(- \log \Pr\limits_\theta{(s_i|p,s_1, \ldots, s_{i-1})} \right)$. Now,
    given a model $\theta$, an inserted canary $c=(p|s)$ where $s$ is an $n$-word sequence, and a set $R$ that consists of $n$-word sequences with each word sampled u.a.r. from the vocabulary, the rank of the canary $c$ can be defined as $\text{rank}_\theta(c; R) = \left|
    \{r' \in R: P_\theta(r' | p) \leq P_\theta(s | p)   \}\right|$. Intuitively, this method captures how strongly the model favors the canary as compared to random chance.
    For our experiments, we consider the size of the comparison set $R$ to be $2 \times 10^6$.
    \item \textbf{Beam Search (BS)} Given a prefix, and the total length of the phrase to be extracted, this method conducts a greedy beam search on a model. As a result, this method functions without the knowledge of the whole canary.
    For our experiments, we use a beam search width of five.
    Using this method, we evaluate if given a 2-word prefix, the canary is among the top-5 most-likely 5-word continuations for the model.
\end{enumerate}

    \mypar{Remark} This experiment is designed to over-estimate what an adversary might be able to learn in a realistic scenario.
    For instance, some of the synthetic users participating in our training process contain number of copies of a canary that is much higher than what would be expected for a user in a practical setting.
    In fact, for any canary with $n_e=200$, the training data of a synthetic user carrying that canary contains 200 copies of the canary.
    Moreover, if $n_u=16$, there are 16 such synthetic users in the training population, each of which participates at a rate 1-2 orders of magnitude higher than any actual device.
    Even for our random sampling method described above, an adversary is assumed to have knowledge of a ``guess" of the canary, and the method provides confidence to the adversary whether the canary was present in the training dataset. For the beam search method, the adversary is assumed to have knowledge of a two-word prefix of the five-word canary, and the method evaluates whether the adversary can extract the canary using a beam search.

\subsection{Empirical Results}
    Table~\ref{table:mem_expts} summarizes the unintended memorization results of a model trained for 2000 rounds using Algorithm~\ref{alg.fedavg} on a training population with actual devices and secret-sharing synthetic devices.

    \begin{table}[ht]
    \centering
    \begin{tabular}{r r@{\hspace{0.3in}} r r r c}
     \toprule
     $n_u$ & $n_e$  & \multicolumn{3}{c}{Random Sampling} & \# canaries found\\
     & & \multicolumn{3}{c}{(approx. rank out of 2M)} & via Beam Search \\
     \midrule
     1 & 1   & 637k, & 1.55M, &1.6M   & 0 / 3 \\
     1 & 14   & 1.6k, &41k, &542k  & 0 / 3 \\
     1 & 200  & 270k, &347k, & 894k  & 0 / 3 \\
     4 & 1  & 281k, &308k, & 1.37M  & 0 / 3  \\
     4 & 14  & 1, &16, & 762  & 1 / 3 \\
     4 & 200  & 263, &904, & 4.9k  & 0 / 3\\
     16 & 1  & 3.7k, &112k, & 129k & 0 / 3 \\
     16 & 14  & 1, & 1, & 1  & 3 / 3  \\
     16 & 200  & 1, & 1, & 1  & 3 / 3  \\
     \bottomrule
    \end{tabular}
    \caption{For each $(n_u, n_e)$ configuration, the approximate rank of the three inserted canaries via Random Sampling, and the number of canaries (final 3 words completed given the first 2) in the top-5 results of Beam Search. The results are for a given prefix length of two.}
    \label{table:mem_expts}
    \end{table}

    First, we observe that all of the inserted canaries having one secret-sharing user (i.e., $n_u = 1$) are far from being memorized, even for the ones when all the examples of the user are replaced by the canary ($n_e = 200$).
    A similar effect can be seen for all the canaries having one insertion per user, even for the ones having 16 users sharing the same canary.
    For four users sharing a canary and having multiple phrases replaced by the canary (i.e., $n_e \in \{4, 200\}$), we observe that almost all of the inserted canaries are nearly memorized as they have very low ranks via the RS method, with one being memorized as it is the most-likely extraction via the BS method.
    Lastly, all of the inserted canaries shared among 16 users, and having multiple phrases replaced by the canary, are memorized as they have a rank one via the RS method, and are extracted via the BS method. It is important to note that the participation rates of our secret-sharing users in training is 1-2 orders of magnitude higher than any of the actual devices. Moreover, learning a phrase used by a sufficient number $n_u$ of users can be desirable; in particular, for large enough $n_u$ this may be necessary to achieve good accuracy as well as the fairness goal of providing good language models to smaller subgroups.

    Thus, our results (Table~\ref{table:mem_expts}) demonstrate that our NWP models trained via DP-FedAvg exhibit very low unintended memorization. In particular, we see canaries start getting memorized when there are $64.4$k occurrences of the canary shared across four users in the training set, whereas they get completely memorized when there are $257.6$k occurrences across 16 users.

    In order to make stronger conclusions, it would be desirable to run several repetitions of our experiment. As indicated in Section~\ref{sec:impl}, running it once involves neural network training spanning three weeks on actual devices with limited computation power. Thus, it is difficult to conduct many repetitions of the experiment.

\section{Practical Considerations for Privacy Guarantees}

\label{sec:prac}
In this section, we delve into some practical considerations to be taken into account while bringing a technique from theory to practice.
\subsection{Proving Differential Privacy Guarantees}
To be able to prove guarantees for Differential Privacy (DP), we formally define the notion here. We first define neighboring datasets (alternatively, training populations in an FL setting). We will refer to a pair of training populations $D,D'$ as neighbors if $D'$ can be obtained by the addition or removal of one user from population $D$.

\begin{defn}[Differential privacy \cite{DMNS06, ODO}] A randomized  algorithm $\cA$ is $(\eps,\delta)$-differentially private if, for any pair of neighboring training populations $D$ and $D'$, and for all events $\cS$ in the output range of $\cA$, we have
$$\Pr[\cA(D)\in \cS] \leq e^{\eps} \cdot \Pr[\cA(D')\in \cS] +\delta$$
where the probability is taken over the random coins of $\cA$.
\label{def:diiffP}
\end{defn}

\mypar{Remark} To relate with the evaluation in Section~\ref{sec:mem}, such a user-level DP guarantee will quantify protection against memorization of any one user's data (i.e., $n_u = 1$). However, extending to the case of $n_u=16$ users (e.g., via a group privacy argument \cite{DR14}) will result in a very weak protection. For instance, a per-user $(1, 10^{-8})$-DP guarantee will result in a guarantee of $(16, 0.53)$-DP for a group of 16 users.

\mypar{Privacy analysis of DP-FedAvg with fixed-size federated rounds (Algorithm~\ref{alg.fedavg})}
Following the analysis of this technique in \cite{DPFGAN19}, the analytical moments accountant \cite{WBK19} can be used to obtain the R\'enyi differential privacy (RDP) guarantee for a federated round of computation that is based on the subsampled Gaussian mechanism, Proposition 1~\cite{mir17} for computing the RDP guarantee of the composition involving all the rounds, and Proposition 3~\cite{mir17} to obtain a DP guarantee from the composed RDP guarantee.

 The analysis above requires several assumptions that require special attention in production FL settings.

\mypar{Sampling uniformly at random}  For the privacy amplification via subsampling~\cite{mir17, WBK19} to apply to Algorithm~\ref{alg.fedavg}, it is required that this sampling be uniformly at random without replacement on each round.

However, in a practical implementation, at any round the server only sees a small subset of the full population. Pace Steering (discussed previously) intentionally limits the number of devices that connect to the server to avoid overloading the system. Further, devices only check-in when they meet availability criteria such as the device being idle, plugged in for charging, and on an unmetered Wi-Fi network. While both of these factors are approximately random, the server cannot precisely characterize this randomness, and can instead only ensure random sampling from the much smaller set of devices that choose to connect. Further, due to dynamic effects introduced by Pace Steering, it is difficult to precisely estimate the total population size.

 \emph{If} we could ensure uniform sampling from a known population size, then upper bounds on $\eps$ and $\delta$ would hold \cite{WBK19} as in Table~\ref{table:epsdelta} . Our best estimate of the actual training population size is $N = 4$M, but for the reasons outlined here, we refrain from making any specific $(\epsilon, \delta)$-DP claims for the training procedure.

  \begin{table}[t]
\centering
\begin{tabular}{c  c}
 \toprule
Device population size $N$ & $\eps \left( \text{for } \delta=N^{-1.1}\right)$\\
 \midrule
 2M & 9.86\\
 3M & 6.73\\
 4M & 5.36\\
 5M & 4.54\\
 10M & 3.27\\
 \bottomrule
\end{tabular}
\caption{Hypothetical upper bounds on $(\eps, \delta)$-DP under the unverifiable-in-production-FL-setting assumptions of a known population size $N$ and uniform sampling. These are computed fixing $\delta = N^{-1.1}$ for the production training described in Section~\ref{sec:prod}, where total rounds  $T=2000$, round participation fraction $q = 20000/N$, and noise standard deviation $\sigma = 3.2\times10^{-5}$.}
\label{table:epsdelta}
\end{table}

\mypar{Secrecy of the sample} Privacy amplification via subsampling requires that the information about which particular users were sampled in any round of training not be accessible to any party other than the trusted central aggregator. This can be challenging to achieve in a distributed setting. However, in addition to all the network traffic being encrypted on the wire, the communication channels between our users and the server are shared for carrying out various other tasks and analytics. Thus, it is difficult for any adversary, even one that is monitoring a communication channel, to confidently draw a conclusion about the participation of a user in our training process.

\subsection{Other Considerations}
Apart from assumptions required for obtaining formal privacy guarantees, there are also few other considerations that need to be made while deploying such a distributed system.
\begin{itemize}
    \item \textbf{Restricted access for user-to-server communication:} For a central DP guarantee in a distributed setting, the updates communicated from each user to the server (trusted central aggregator) should be accessible only by the server. To ensure this, all network traffic is encrypted on the wire in the framework~\cite{FL19} our implementation uses. This includes any communication from the users to the server and vice-versa.

    \item \textbf{Privacy cost of hyperparameter tuning:} Prior work~\cite{GLMRT10, CMS11, CV13, BassilyST14, DPDL, LT19} has shown that hyperparameter tuning using sensitive data can incur a significant privacy cost. Thus, we perform extensive experiments for tuning various hyperparameters in our technique using publicly-available language datasets so as to not affect the privacy of any user participating in our training process.
\end{itemize}

\section{Conclusions}
\label{sec:concl}

This work details the first production next-word prediction (NWP) model trained using on-device data while leveraging the Differentially Private Federated Averaging technique, and an existing FL infrastructure.
We show that our trained NWP model has superior utility than the existing baseline.
Using an end-to-end measurement process, we also empirically demonstrate the remarkable extent to which models trained via our implementation are able to withstand unintended memorization.
Lastly, we shed light on some of the considerations to be made for bringing such a technique from theory to a real-world implementation.
Keeping practical considerations in mind, a potential novel direction to strengthen the privacy guarantees of such a system is to incorporate techniques like random check-ins \cite{BKMTT} into the training framework. We leave this for future work.

\ifsp
\else
\section*{Acknowledgements}
\label{sec:acks}
The authors would like to specially thank Peter Kairouz, Ananda Theertha Suresh, Kunal Talwar, Abhradeep Thakurta, and our colleagues in Google Research for their helpful support of this work, and comments towards improving the paper.
\fi

\bibliographystyle{alpha}
\bibliography{references}
\appendix
\section*{Ablation Studies}
\label{sec:abl}
Now, we present the results of the ablation studies on using DP-FedAvg on the Stack Overflow dataset. Table~\ref{table:optimizer_ablation} shows results from an ablation study on the effect of server optimizer parameters. We observe that using Nestorov momentum works better than SGD and Adam.

\begin{table}[ht]
    \centering
    \begin{tabular}{ l  c }
     \toprule
     Server Optimizer params &  Top-1 Recall [\%] \\
     \midrule
     Adam, $\eta_s=1\times10^{-5}$ & 4.73 \\
     Adam, $\eta_s=5\times10^{-5}$ & 15.49 \\
     Adam, $\eta_s=1\times10^{-4}$ & 19.78 \\
     Adam, $\eta_s=2\times10^{-4}$ & 21.92 \\
     Adam, $\eta_s=5\times10^{-4}$ & 23.38 \\
     \midrule
     Momentum, $\eta_s=0.5, \mu=0.9$ & 23.03 \\
     Momentum, $\eta_s=1.0, \mu=0.9$ & 23.69 \\
     Momentum, $\eta_s=0.5, \mu=0.99$ & 24.16 \\
     \textbf{Momentum}, \boldmath$\eta_s=1.0, \mu=0.99$ & \textbf{24.15} \\
     \midrule
     SGD, $\eta_s=0.5$ & 18.49 \\
     SGD, $\eta_s=0.7$ & 19.52 \\
     SGD, $\eta_s=1.0$ & 20.41 \\
     \bottomrule
    \end{tabular}
    \caption{Ablation study on server optimizer parameters. Hyperparameters used for training the model on real devices are highlighted in bold.}
    \label{table:optimizer_ablation}
\end{table}

Table~\ref{table:batchsize_ablation} shows results from another ablation study on the effect of various batch sizes and learning rates on the client. Batch sizes and learning rates on the client don't seem to have a large impact on performance, with batch sizes from $|b| = 5$ to $|b| = 50$ demonstrating similar performance.

\begin{table}[ht]
    \centering
    \begin{tabular}{l c }
     \toprule
     Client optimizer params &  Top-1 recall [\%] \\
     \midrule
     $|b|=5, \eta_c=0.1$ & 23.92 \\
     $|b|=5, \eta_c=0.5$ & 24.03 \\
     $|b|=10, \eta_c=0.2$ & 24.03 \\
     $|b|=10, \eta_c=0.5$ & 23.96 \\
     $|b|=20, \eta_c=0.3$ & 24.00 \\
     $|b|=20, \eta_c=0.5$ & 24.03 \\
     \boldmath$|b|=50, \eta_c=0.5$ & \textbf{24.15} \\
     \bottomrule
    \end{tabular}
    \caption{Ablation study on client optimizer parameters. Hyperparameters used for training the model on real devices are highlighted in bold.}
    \label{table:batchsize_ablation}
\end{table}

\begin{figure}[ht]
	\centering
    \includegraphics[width=\linewidth]{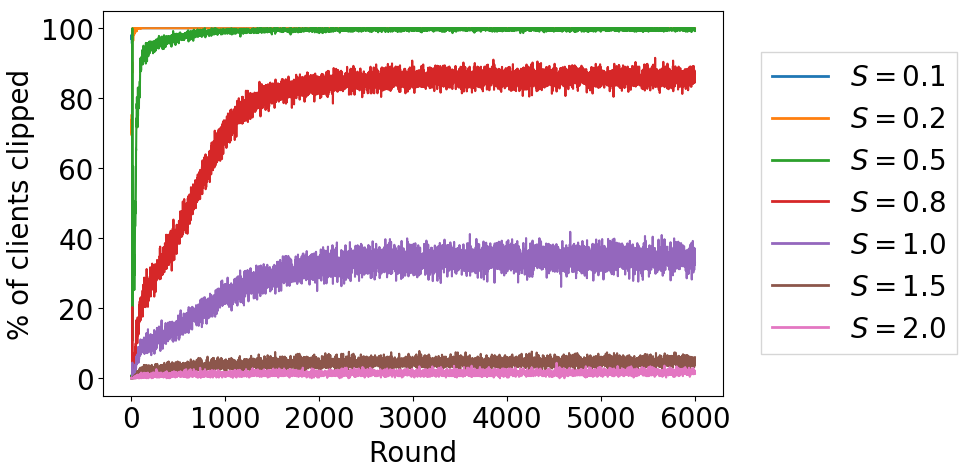}
	\caption{\% of clients clipped vs. round for different values of clipping norm ($S$).}
	\label{fig:clipping_ablation_plot}
\end{figure}

Table~\ref{table:clipping_ablation} shows results from an ablation study on various clipping values used for clipping the user updates. Figure~\ref{fig:clipping_ablation_plot} shows the percentage of clients clipped across the duration of training, for different values of the clipping norm. We observe that clipping a large fraction of clients works better. Below a certain value ($S=0.2$ in this case), almost all the clients get clipped, and further clipping is equivalent to decreasing the server learning rate.

\begin{table}[ht]
    \centering
    \begin{tabular}{l c}
     \toprule
     Clipping norm &  Top-1 recall [\%]  \\
     \midrule
     $S=0.1$ & 23.78 \\
     $S=0.2$ & 23.97 \\
     $S=0.5$ & 24.09 \\
     \boldmath$S=0.8$ & \textbf{24.15} \\
     $S=1.0$ & 24.12 \\
     $S=1.5$ & 23.81 \\
     $S=2.0$ & 23.45 \\
     \bottomrule
    \end{tabular}
    \caption{Ablation study on clipping norm values. Hyperparameters used for training the model on real devices are highlighted in bold.}
    \label{table:clipping_ablation}
\end{table}

These ablation studies are not meant to serve as an extensive sweep of the hyperparameters. These are presented demonstrate that it's feasible to tune hyperparameters for DP-FedAvg on a public corpus and avoid incurring any additional privacy cost.

\end{document}